\documentclass[a4paper,conference]{IEEEtran}
\ifCLASSINFOpdf
\else
\fi

\usepackage{graphicx}
\usepackage{amsmath}
\usepackage{booktabs}
\usepackage{subfigure}
\usepackage{xcolor}
\begin{document}
\title{ContinuityLearner: Geometric Continuity Feature Learning for Lane Segmentation}

\author{\IEEEauthorblockN{Haoyu Fang}
\IEEEauthorblockA{NYU Tandon School of Engineering\\
New York, NY 11201\\
Email: haoyu.fang@nyu.edu}
\and
\IEEEauthorblockN{Jing Zhu}
\IEEEauthorblockA{NYU Tandon School of Engineering\\
New York, NY 11201\\
Email: jingzhu@nyu.edu}
\and
\IEEEauthorblockN{Yi Fang}
\IEEEauthorblockA{ECE Department\\
New York University Abu Dhabi\\
Email: yfang@nyu.edu}}


%


\maketitle

\begin{abstract}
Lane segmentation is a challenging issue in the autonomous driving system designing because lane marks show weak textural consistency due to occlusion or extreme illumination but strong geometric continuity in traffic images, from which general convolution neural networks (CNNs) are not capable of learning semantic objects. To empower conventional CNNs in learning geometric clues of lanes, we propose a deep network named ContinuityLearner to better learn geometric prior within lane. Specifically, our proposed CNN-based paradigm involves a novel Context-encoding image feature learning network to generate class-dependent image feature maps and a new encoding layer to exploit the geometric continuity feature representation by fusing both spatial and visual information of lane together. The ContinuityLearner, performing on the geometric continuity feature of lanes, is trained to directly predict the lane in traffic scenarios with integrated and continuous instance semantic. The experimental results on the CULane dataset and the Tusimple benchmark demonstrate that our ContinuityLearner has superior performance over other state-of-the-art techniques in lane segmentation.
\end{abstract}

\section{Introduction}
\begin{figure}
  \includegraphics[width=1\linewidth]{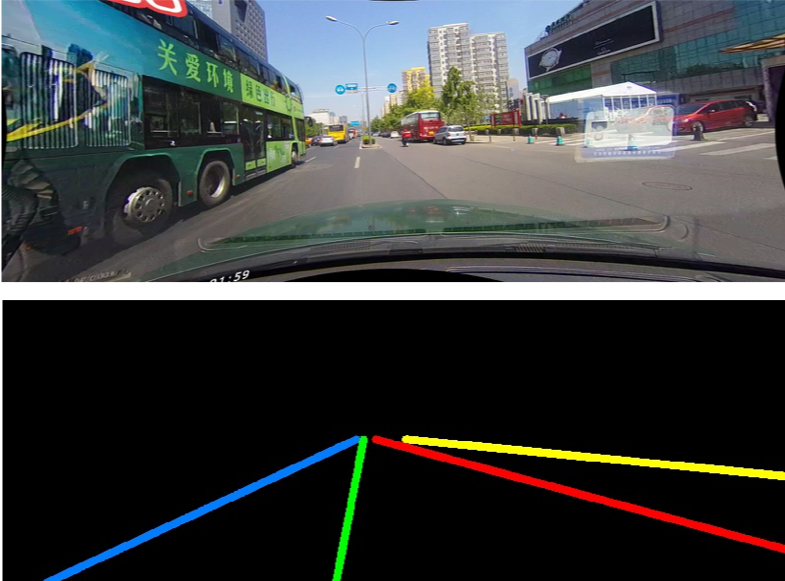}\hfill
  \caption{An example of lane segmentation on CULane Dataset \cite{pan2018SCNN}. The fist row is a traffic image and the second is corresponding annotations of the lane boundaries. The example illustrates four common challenging cases that may disable a deep learning network to segment perfect lane boundaries in practical applications. From left to right, the challenges are occlusion, noncontinuous visual information (dash line), no sufficient visual clues (no line) and strong interference (optical reflection).}
  \label{fig1}
\end{figure}

Autonomous vehicles and driving assistant systems have been a major focus of the computer vision research community. One of the key components of autonomous driving is to perceive and understand the surrounding environment of the vehicle. In particular, environment comprehension tasks, i.e. lane detection, semantic segmentation, become essential as they assist the self-driving car to locate itself between lane boundaries. Despite years of research, lane segmentation is still a challenging task due to several reasons (shown in Fig. \ref{fig1}): 1) complex traffic scenarios (e.g. self-occlusion and vehicle-occlusion) reduce visual information for a vision-based algorithm to recognize lanes; 2) various lane patterns, especially those lack essential visual features, bring great difficulties to learn distinctive feature of road lanes; 3) environmental interference like low illumination, reflection brings unpredictable influence on captured images even in the same scenes.

Before neural networks prevailing in computer vision tasks, traditional methods to segment lane boundaries are almost based on handcrafted features, which can only work in limited scenarios under strict constraints such as parallel lanes \cite{aly2008real,deusch2012random,jiang2010computer,nieto2008robust} or close to straight lanes \cite{li2015lane,niu2016robust}. The handcrafted features, such as color \cite{chiu2005lane}, bar filter \cite{teng2010real}  and structure tensor \cite{loose2009kalman} etc, usually are not robust enough to handle all the challenging cases of diverse real-world road scenarios. With neural networks, especially fast development of deep-learning based approaches, works on deep learnable feature instead of handcrafted features have become a new promising research direction. 

Although CNNs have shown impressive performance in lane segmentation, they exploit mainly on visual information that might be insufficient in real-world traffic scene. This challenge greatly limits the networks to provide a satisfactory performance in practice. However, unlike many other traffic objects, road lanes are observed to have strong structure prior but less appearance clues. Exploitation on this structure prior or geometric feature of lanes that has long continuous shape, shows importance on performance improvement. Besides, geometric feature is not only conveyed by visual knowledge but also contained in lane pixel locations (coordinates of pixels that belongs to road lanes), which provides distinctive feature even if visual knowledge is not perfect or sufficient. Triggered by the motivation of cultivating geometric feature, we propose a novel deep CNN named ContinuityLearner, which is the pioneering approach to directly learn structural feature beneath lane-pixel locations and exploit visual information of lane at the same time. The proposed ContinuityLearner, performing on a novel geometric continuous feature representation that fuses visual feature and pixel locations together, predicts lanes in traffic scenarios with integrated and smooth instance semantics. Therefore, the major contributions of the paper are as follows:

\begin{itemize}
\item We firstly propose a lane segmentation paradigm that applies geometric continuity prior of lanes' shape as a global constraint to refine the CNN-based semantic. The geometric continuity feature is a robust and comprehensive representation for networks to learn the semantic of lanes in the traffic image. 
\item We specifically develop a \textit{Context-encoding Image  Feature  Learning  Network} to encode class-dependent image featuremaps for various lanes in the image. 
\item We propose a novel \textit{Geometric Continuity Feature Encoding Layer} that fuses the class-dependent image with pixels' locations to jointly exploit textual knowledge of lanes as well as their spatial information.
\item Our proposed model outperforms several state-of-the-art methods in experiment conducted on Tusimple and CULane dataset, especially in the cases where visual (textual) knowledge of lanes is missing due to occlusion and extreme illumination.
\end{itemize} 

\section{Related works}
Development of deep learning techniques has stimulated a promising research direction for lane detection. \cite{li2015lane} uses a spiking neural network to first extract information about lane edges and then employs Hough transform for lane detection. However, this method is susceptible to occluded lane boundaries. \cite{he2016accurate} using the front-view and top-view of the input image, proposes DVCNN (Dual-View Convolutional Neural Network) for detecting lanes. \cite{gurghian2016deeplanes} proposes a deep neural network namely DeepLanes that estimates the position of the lanes. \cite{li2017deep} develops a multi-task CNN network that extracts geometric information about the lanes and a Recurrent Neural Network (RNN) based model that detects lane boundaries. \cite{neven2018towards} adopts an instance segmentation approach to lane detection in which the proposed end-to-end network outputs an instance map. Note that each lane pixel in the instance map indicates which lane it belongs to. In this paper, in order to take into account changes in road plane, a unique learned transformation (outputted by H-Net) is employed instead of using the traditional fixed "bird's-eye view" transformation. \cite{wang2018lanenet} proposes LaneNet which divides the lane detection task into two steps: lane edge proposal and lane line localization. In the first stage, a lane edge proposal network classifies each pixel of the input image and outputs lane edge proposals which contain information about the edges of the lanes. These lane proposals are then fed to the lane line localization networks to finally output the detected lanes. \cite{pan2018spatial} proposes a Spatial CNN which employs layer-by-layer convolutions to learn the spatial information and detect lane lines.

Some other recent works, using global geometric constraints, have been conducted in the realm of lane detection. One geometric feature that has been considered important in lane detection is vanishing point \cite{su2018vanishing,lee2017vpgnet}.  Lee et al. propose VPGNet \cite{lee2017vpgnet}, a network that not only detects different types of lane lines but also detects road markings under unfavourable weather conditions. \cite{zhang2018geometric} proposes a multi-task model for lane segmentation and lane boundary detection.
\begin{figure*}[ht]
    \centering
    \includegraphics[scale=0.45]{ 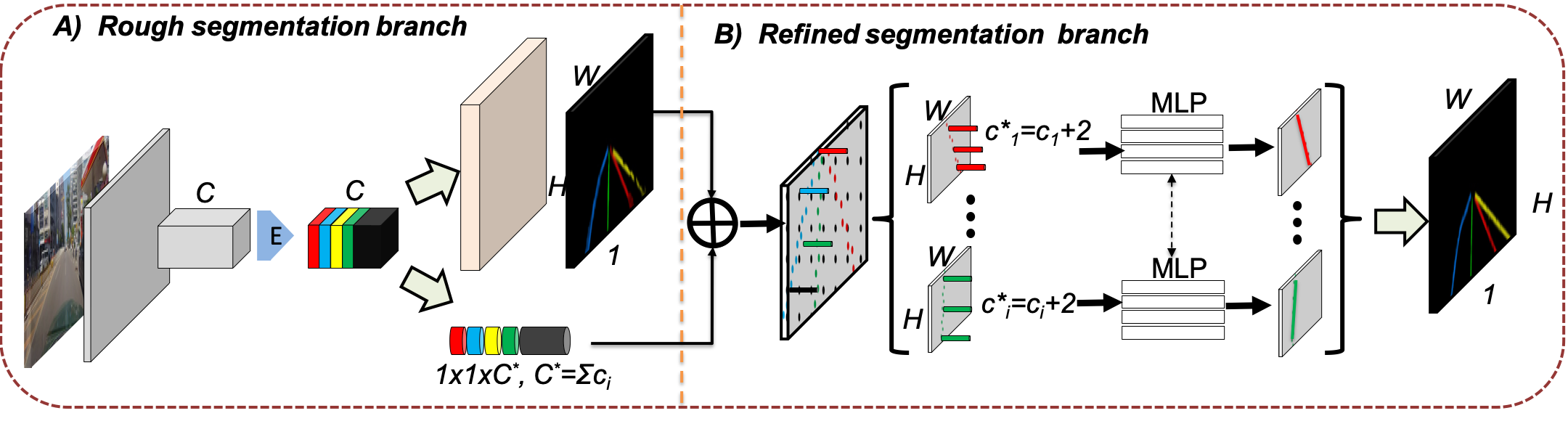}
    \caption{Pipeline of ContinuityLearner which consists two branch: A) the rough segmentation framework based on a context-encode image feature learning network; B) the refined segmentation framework that applies a geometric continuity feature encoding layer and a PointNet-based \cite{pointnet} network. Note that W, H and C denote width and height and channels of feature maps respectively. $d_i$ refers to the channels of feature of object $i$. $2$ in (B) represents the location of pixel in image plane (x,y). }
    \label{pipeline}
\end{figure*}

\section{Method}
The proposed model (depicted in Fig.\ref{pipeline}) consists of two branch: A) the rough segmentation branch and B) the refined segmentation branch. The rough segmentation branch is mainly formed by a context-encoding  image  feature  learning  network that captures the feature statistics as a global semantic context and generates a coarse semantic and class-dependent feature maps. The refined segmentation branch involves a geometric continuity feature encoding layer which converts the rough semantic into a 2D point cloud and then fuses the point set conveying the spatial information with the image feature learned in (A). The multilayer perceptron (MLP)-based network in the branch (B) predicts possible missing key lane points to generate a more integrated and continuous semantic of road lanes.

\subsection{Rough Segmentation Branch}
\begin{figure}[ht]
    \centering
    \includegraphics[scale=0.35]{ 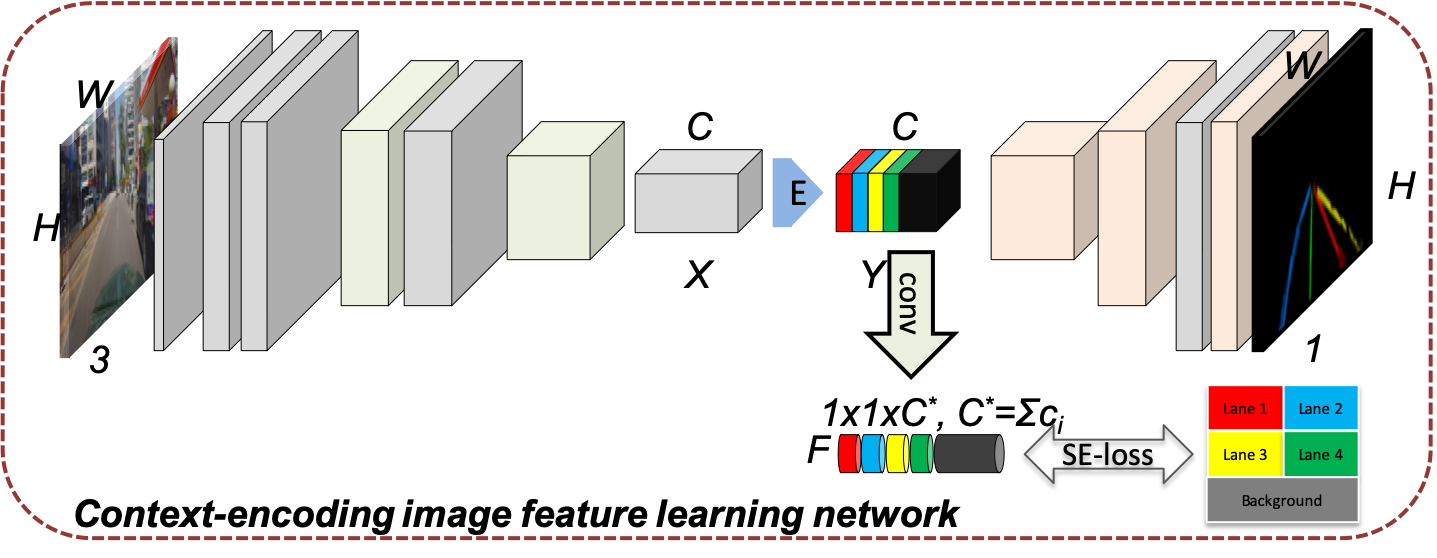}
    \caption{The context-encoding image feature learning network}
    \label{rough}
\end{figure}

The rough segmentation branch in Fig.\ref{rough}, fed with a color traffic image, generates a coarse semantic of lanes and class-dependent feature maps which encode rich information what lane objects are in the input image. This branch is mainly formed by context-encoding feature learning network that predicts a set of attention weights to highlight the class-dependent feature maps.

\noindent\textbf{Context-encoding module} The proposed context-encoding feature learning network takes advantages of an encoding module \cite{context_encode2} that learns a dictionary to describe the semantic context of dataset and generates attention weights with rich contextual knowledge. The encoder module is fed with an input feature map with the shape $C \times H \times W$ as a set of C-dimensional input feature $X = \{x_1,...,x_N\}$, where $N$ is total number of features given by $H \times W$.
The module learns an inherent $D = \{d_1,....,d_k\}$ containing $K$ number of codewords and a set of smoothing factor of visual centers $S=\{s_1,...,s_k\}$. The context-encoding network outputs the residual encoder by aggregating the residuals with soft-assignment weights $e_k = \Sigma_{i=1}^{N}e_{ik}$, where
\begin{equation}
\centering
e_{ik} = \frac{exp(-s_k||x_i-d_k||^2)}{\Sigma_{j=1}^{K}exp(-s_j||x_i-d_j||^2)}(x_i-d_k).
\label{eq3.1}
\end{equation}
We aggregate the encoders by $e=\Sigma_{k=1}^K\phi(e_k)$, where $\phi$ denotes Batch Normalization with ReLU activation.

\noindent\textbf{Class-dependent feature maps}
The utility of the encoders $e$ reduces the dimension of the feature representations and we produce an attention weights which highlight the independence of different lane objects in feature dimension by $A=\delta(We)$, where $W$ represents convolution layers' weights and $\delta$ denotes an activation function (i.e. the sigmoid function in this paper). The encoded image feature denoted by $Y$ is calculated by
\begin{equation}
\centering
Y = X \otimes A,
\label{eq3.2}
\end{equation}
where $\otimes$ is a channel-wise multiplication and $Y$ has the same dimension as the input feature map $X$. To further fit the encoded image feature map into a 2D point cloud format, a convolution layer denoted by $conv(\cdot):R^{C \times H \times W} \xrightarrow{} R^{C^* \times 1 \times 1}$ is employed to extract a class-dependent feature, where $C^* = \Sigma_M c_i$ and $M$ is the number of detected lanes. $c_i$ represents the feature that encodes image feature of the $i$th lane.
\begin{equation}
\centering
F = conv(Y),
\label{eq3.3}
\end{equation}

In the training process, a semantic encoding loss (SE-loss) \cite{context_encode2} that makes individual predictions for the presences of object categories in the scene and learns with binary cross entropy loss is applied to force the network to understand the global semantic information and calculate the class-dependent feature maps according to the lane annotations (e.g. lane No.1, lane No.2 and etc.). 

\subsection{Refined Segmentation Branch}
Although lanes demonstrate relatively weak textual coherence due to the lack of visual information, they have stronger geometric consistency in the image. The geometric continuity of lanes is reflected by the location of lane pixel in the image plane. Therefore, the proposed refined segmentation branch concentrates on exploiting spatial knowledge of lane pixels to construct the geometric feature representation of lanes. This module converts the coarse semantic into several incomplete 2D point clouds and appends class-dependent image feature maps with points' location in the corresponding point set.

Before converting, a filtering process with a fixed probability threshold will perform on the rough segmentation to guarantee that only prediction with high confidence will be regraded as lane pixels and further reduce the false positive prediction in the rough segmentation. Then a MLP-based network \cite{pointnet} is trained to fill possible key points into the incomplete point set to reconstruct a continuous prediction which better reflects the geometric shape of lanes. 

\begin{figure}[ht]
    \centering
    \includegraphics[scale=0.5]{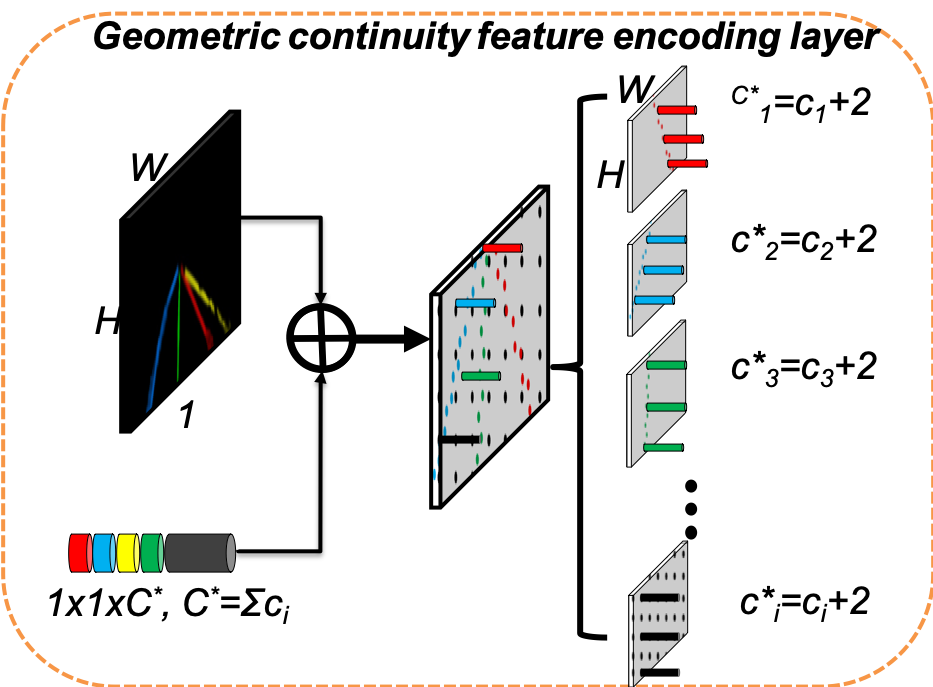}
    \caption{Geometric continuity feature encoding layer. }
    \label{geo}
\end{figure}

\noindent\textbf{Geometric feature encoding layer} This module generates a geometric continuity feature representation by concatenating the class-dependent image feature of a specific lane with locations of all its lane pixels (in Fig. \ref{geo}). We convert the rough segmentation results $I_{rough} \in R^{W \times H}$ with lane instances' semantic into a 2D point cloud $P \in R^{N*2}$ by regarding each lane pixel in the coarse segmentation as an isolated point in the image plane, where $N$ is the total number of points. A specific order is taken to extract the points (i.e. top to bottom and left to right in this paper) and the point cloud contains all lanes' pixels in the rough segmentation $P = \{P_i|i = 1,2,\cdot \cdot \cdot,M\}$, where $M$ denotes the number of predicted lanes in the rough segmentation. For each points of $i$th lane ($P_i$), the geometric continuity feature representation $G_i$ is produced by concatenating the point's location with object's independent image feature $c_i$ and the point set $P_i$ is encoded into the high dimensional feature space by $P_i \in R^{{M_i} \times 2} \xrightarrow{} G_i \in R^{{M_i} \times (2+c_i)}$. 

\noindent\textbf{Point set completion}
Considering the segmentation of an arbitrary lane as a point set. The missing areas in rough segmentation of the lane result in some key points that reflect the shape of the lane missing in the corresponding point set. we use a MLP-based network \cite{pointnet} to predict those missing points and reconstruct a completed point set which better represents the semantic of lanes. The MLP-based network denoted by $MLP(\cdot)$ takes the geometric continuity feature $G_i$ as input and predicts missing points of the incomplete set by
\begin{equation}
\centering
P^*_i = MLP(G_i),
\label{eq3.4}
\end{equation}
where $P^*_i \in R^{M_i^*\times 2}$ is the completed point set and $M_i^* \geq M_i$. Finally, the completed point set is converted into image pixels by a quantization process $h: P^*_i \in R^{M_i^*\times 2}\xrightarrow{} P^{**}_i \in Z^{M_i^*\times 2}$. Those pixels are finally deployed in image plane and form the lane semantics.

\section{Experiments and Results}
We test the proposed ContinuityLearner on the Tusimple dataset \cite{tuSimple} and the challenging CULane dataset \cite{pan2018SCNN}, which are both large scale datasets containing sufficient images for training, validation and testing in various traffic conditions. The experiments will be analyzed in this section in details. To conduct fair comparison, we employ the same evaluation metrics as \cite{pan2018SCNN} and \cite{endtoend} to test performance on CULane and Tusimple testing sets respectively.

\begin{figure*}[htbp]
\centering

\subfigure[Input Images]{
    \begin{minipage}[t]{0.23\linewidth}
        \centering
        \includegraphics[width=1\linewidth]{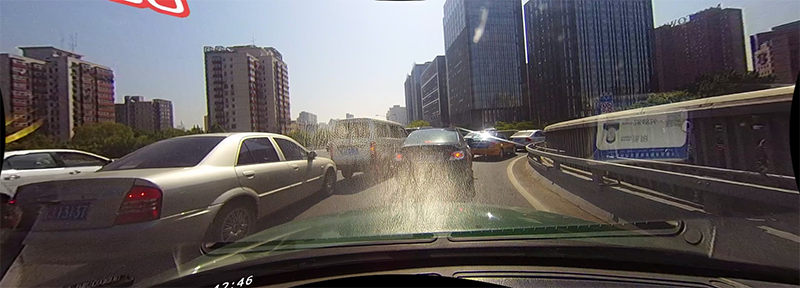}\\
        \vspace{0.02cm}
        \includegraphics[width=1\linewidth]{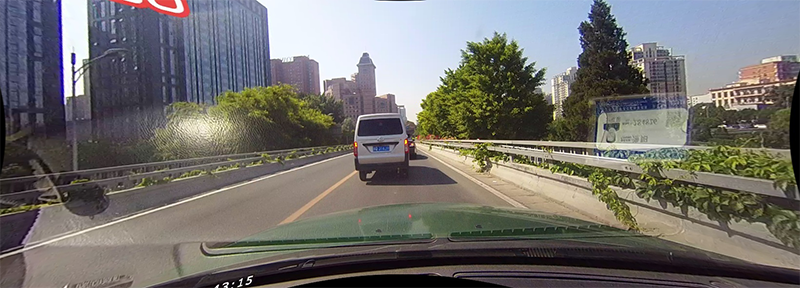}\\
        \vspace{0.02cm}
        \includegraphics[width=1\linewidth]{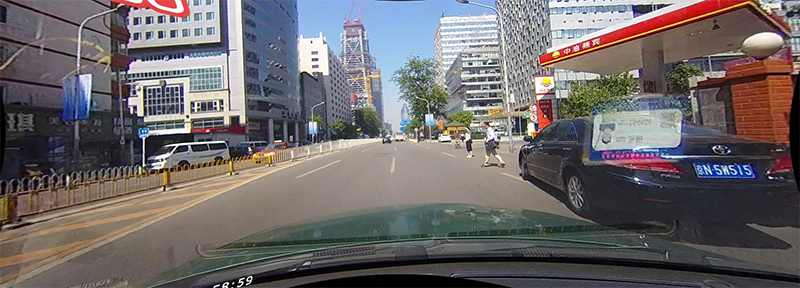}\\
        \vspace{0.02cm}
        \includegraphics[width=1\linewidth]{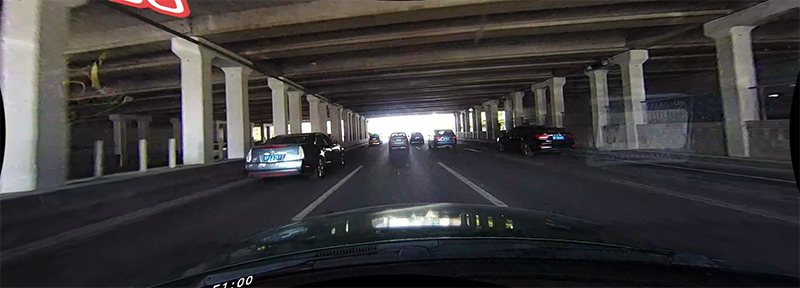}\\
        \vspace{0.02cm}
        \includegraphics[width=1\linewidth]{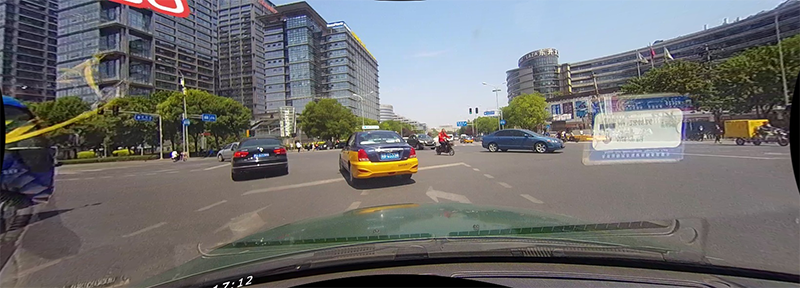}\\
        \vspace{0.01cm}
    \end{minipage}
}
\subfigure[ResNet \cite{ResNet}]{
    \begin{minipage}[t]{0.23\linewidth}
        \centering
        \includegraphics[width=1\linewidth]{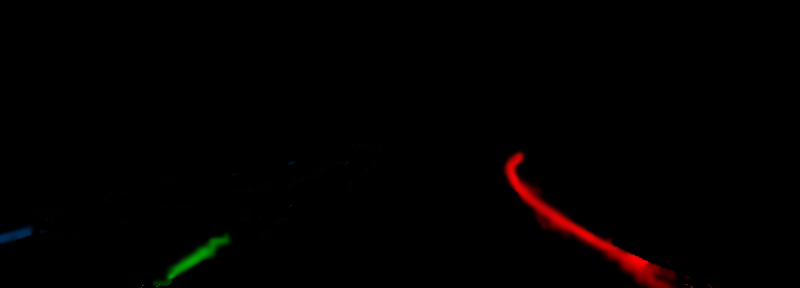}\\
        \vspace{0.02cm}
        \includegraphics[width=1\linewidth]{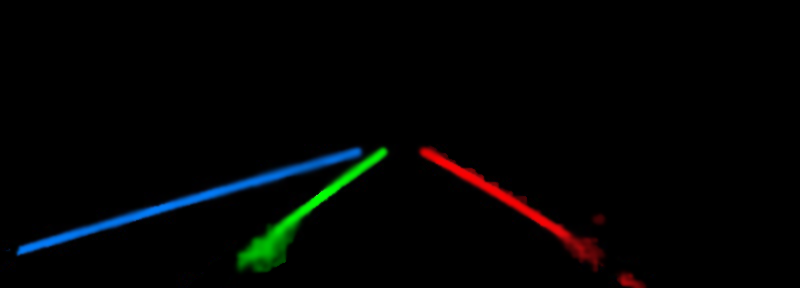}\\
        \vspace{0.02cm}
        \includegraphics[width=1\linewidth]{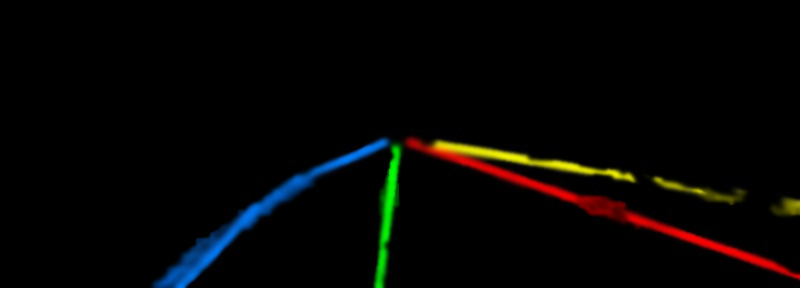}\\
        \vspace{0.02cm}
        \includegraphics[width=1\linewidth]{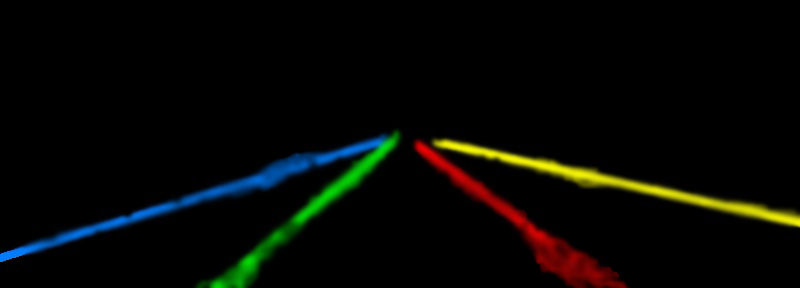}\\
        \vspace{0.02cm}
        \includegraphics[width=1\linewidth]{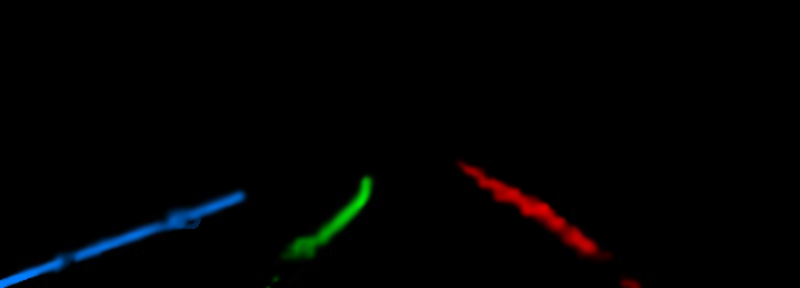}\\
        \vspace{0.01cm}
    \end{minipage}
}
\subfigure[Our Results]{
    \begin{minipage}[t]{0.23\linewidth}
        \centering
        \includegraphics[width=1\linewidth]{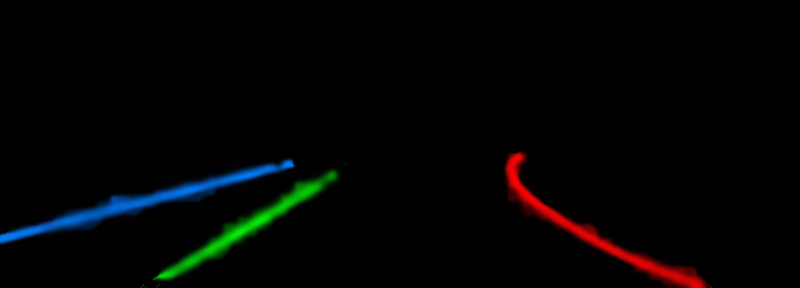}\\
        \vspace{0.02cm}
        \includegraphics[width=1\linewidth]{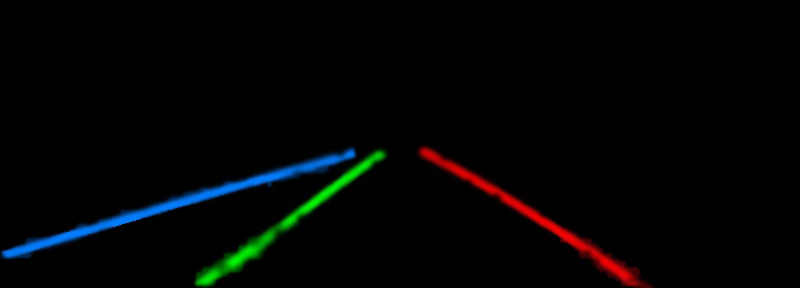}\\
        \vspace{0.02cm}
        \includegraphics[width=1\linewidth]{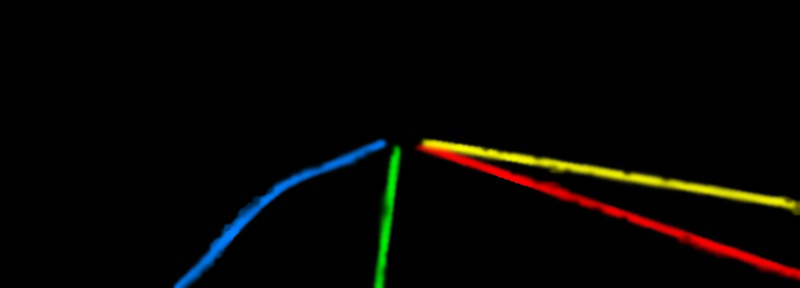}\\
        \vspace{0.02cm}
        \includegraphics[width=1\linewidth]{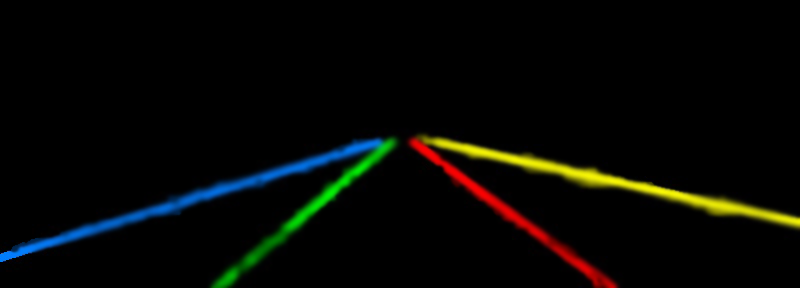}\\
        \vspace{0.02cm}
        \includegraphics[width=1\linewidth]{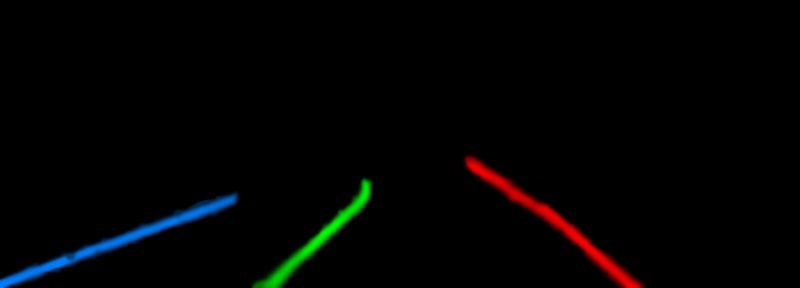}\\
        \vspace{0.01cm}
    \end{minipage}
}
\subfigure[Ground Truth]{
    \begin{minipage}[t]{0.23\linewidth}
        \centering
        \includegraphics[width=1\linewidth]{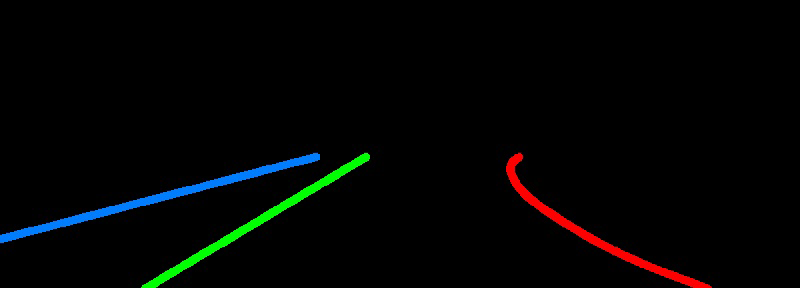}\\
        \vspace{0.02cm}
        \includegraphics[width=1\linewidth]{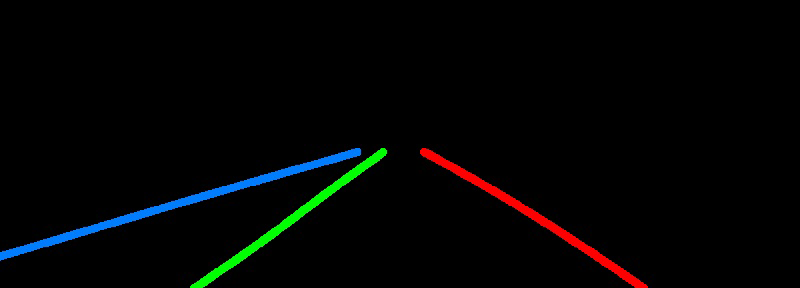}\\
        \vspace{0.02cm}
        \includegraphics[width=1\linewidth]{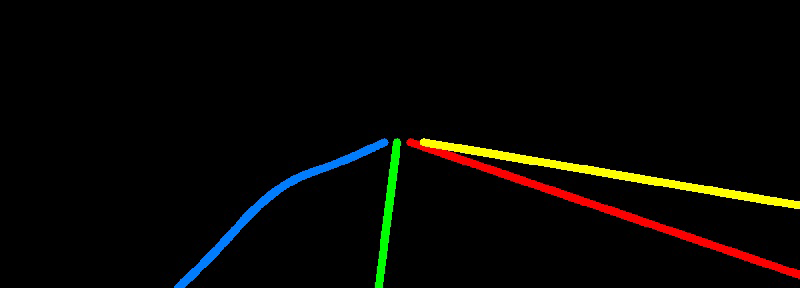}\\
        \vspace{0.02cm}
        \includegraphics[width=1\linewidth]{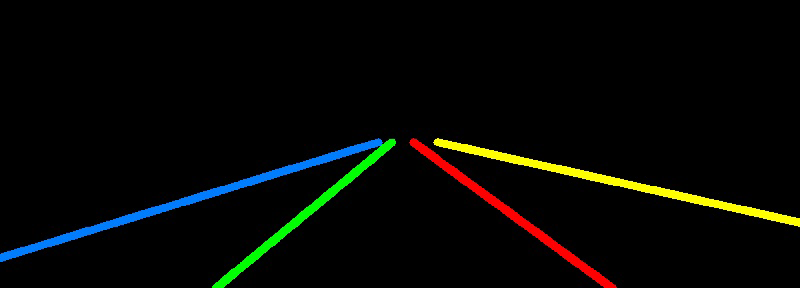}\\
        \vspace{0.02cm}
        \includegraphics[width=1\linewidth]{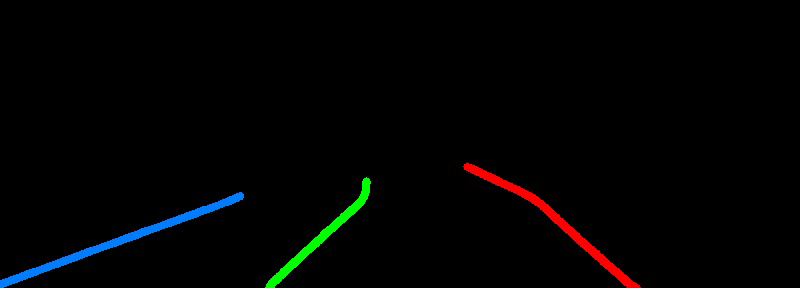}\\
        \vspace{0.01cm}
    \end{minipage}
}
\centering
\par
\caption{Experimental results on CULane dataset \cite{pan2018SCNN}. The first column is the input image, the second is segmentation results by ResNet101 \cite{ResNet}, the third column is our results and the final column is the ground truth. In each sub-figure, the three or four lanes are marked in different colors (blue, green, red and yellow from left to right).}
\vspace{-0.1cm}
\label{fig6}
\end{figure*}

We train the ContinuityLearner on the Tusimple dataset. The model are trained for 2300 episodes with batch size 8 using a standard SGD optimizer with initial learn rate 0.01, momentum 0.9. The learning rate gets halved every other epoch. To save computation and memory usage, we resize the images of TuSimple to $368 \times 640$. For CULane experiments, we use identical hyper-parameters to the former settings but we train the model on the full CULane training dataset for 60K episodes and resize the image of CULane to $288 \times 800$. Moreover, we set the probability threshold of the filter performing on rough segmentation is $70\%$ (regarding prediction with $70\%$ or higher confidence scores as lane pixels) for all experiments.

A quantitative comparison between the proposed method and several state-of-the-art (SOTA) approaches are given in Table \ref{tusimple_results}. While the ContinuityLearner demonstrates the SOTA level instance segmentation accuracy, it also provides the lowest $FN$ rate among all advanced approaches. The ContinuityLearner learns the geometric feature of lanes so that it is still robust to recognise lanes even in dash lane or no lane cases that lack textual clues and confuse conventional algorithms which focus on exploiting visual information only (i.e. SCNN \cite{pan2018SCNN} and FastDraw \cite{FastDraw} etc.). Similarly, another quantitative comparison on CULane \cite{pan2018SCNN} is provided in Table \ref{culane_results}, which clearly illustrates the robust performance of the proposed method in various traffic scenes. In the Table \ref{culane_results}, the proposed method achieves the highest F1 score in three most challenging traffic scenes due to lack of visual information of lanes (i.e. crowded, no lines and shadow traffic scenes.) and also outperforms the SOTA methods in the average score.

In the qualitative comparison, Fig. \ref{fig6} shows the robust performance of the proposed method in instance segmentation of various lanes in CULane testing dataset. We can observe that the proposed network segments different numbers of lanes in various shapes (curve, dash or straight lines). The first row illustrates the instance segmentation in a dazzle traffic scene. Three instances of lane with one in curve shape are segmented more smoothly than the ResNet-101's results. In third row, we observe that the proposed network accurately predicts the semantics of a curve lane and a dash line. In the same row, the proposed method accurately predicts a lane (in blue) which is occluded by a vehicle and has no visual information at most part of the image. The fourth row illustrates the ContinuityLearner's performance is not influenced by the illumination condition and predicts all lanes in the image.
\begin{table}[]
\centering
\caption{\label{tusimple_results}Comparison of experimental results of ContinuityLearner with other state-of-the-art methods on the Tusimple dataset \cite{tuSimple}. $Acc$, $FP$ and $FN$ denotes accuracy, false positive score and false negative score respectively. Higher $Acc$ but lower $FP$ and $FN$ represent better performance.}
\begin{tabular}{@{}llll@{}}
\toprule
\textbf{Approach} &\textbf{Acc (\%)} &\textbf{FP} &\textbf{FN}\\
\hline
M. Ghafoorian et al. \cite{EL-GAN}   &94.9 &0.059 &0.067\\
J. Philion et al. \cite{FastDraw}  &95.2 &0.076 &0.045\\
D. Neven et al. \cite{endtoend}&\textbf{96.9} &\textbf{0.044} &0.020\\ 
D. Neven et al. \cite{endtoend}   &96.5 &0.085 &0.027\\
D. Neven et al. \cite{endtoend}  &96.2 &0.236 &0.036\\
P. L et al. \cite{pan2018SCNN}     &96.5 &0.062 &0.018\\
Z. Wang et al. \cite{wang2018lanenet}  &96.4 &0.078 &0.024\\
Y. Hou et al. \cite{distillation} &96.6 &0.060 &0.021\\
\hline
\textbf{ContinuityLearner} &96.2 &0.069 &\textbf{0.016} \\\bottomrule
\end{tabular}
\label{tab0}
\end{table}

\begin{table*}[]
\centering
\caption{\label{culane_results}Comparison of experimental results of ContinuityLearner with other state-of-the-art methods on the CULane \cite{pan2018SCNN} test set. Harmonic mean (F1 measure with an IoU threshold of 0.5) are reported. Higher F1 is better.}
\begin{tabular}{@{}llllllllll@{}}
\toprule
\textbf{Approach} &\textbf{Normal} &\textbf{Crowded}  &\textbf{Night} &\textbf{No lines} &\textbf{Shadow}   &\textbf{Arrow} &\textbf{Dazzle} &\textbf{Curved} &\textbf{Total}\\ \midrule
F. Vision et al. \cite{ReNet}    &83.3 &60.5 &56.3 &34.5 &55.0 &74.1 &48.2 &59.9 &62.9\\
Krahenbuhl et al. \cite{DenseCRF}  &81.3 &58.8 &54.2 &31.9 &56.3 &71.2 &46.2 &57.8 &61.0\\
C. Szegedy et al. \cite{ResNet}  &87.4 &64.1 &60.6 &38.1 &60.7 &79.0 &54.1 &59.8 &66.7\\
P.L. et al. \cite{pan2018SCNN}      &90.6 &69.7 &66.1 &43.4 &66.9 &84.1 &58.5 &64.4 &71.6\\
J. Phillion et al.\cite{FastDraw} &85.9 &63.6 &57.8 &40.6 &59.9 &79.4 &57.0 &65.2 &-\\
Y. Hou et al. \cite{distillation} &\textbf{90.7} &70.0 &\textbf{66.3} &43.5 &67.0 &\textbf{84.4} &\textbf{59.9} &\textbf{65.7} &71.8\\
\hline
Base model (ResNet-101)&90.2 &68.2 &65.9 &41.7 &64.6 &84.0 &59.8 &65.5 &70.8\\
\textbf{ContinuityLearner} &90.5 &\textbf{70.3} &66.2 &\textbf{43.9} &\textbf{67.2} &82.3 &58.8 &64.0 &\textbf{72.1}\\ \bottomrule
\end{tabular}
\label{tab1}
\end{table*}

\begin{table*}[]
\centering
\caption{\label{tab:table-name}Ablation study on the CULane test set. F1 scores (with an IoU threshold of 0.5) are reported. The proposed ContinuityLearner equals to the aggregation of the base model and A) the rough segmentation branch and B) the refined segmentation branch. The blue texts represent the best score based on ResNet-50 as the base model and the red represent the best based on ResNet-101.}
\begin{tabular}{@{}llllllllll@{}}
\toprule
\hline
\textbf{Approach} &\textbf{Normal} &\textbf{Crowded}  &\textbf{Night} &\textbf{No lines} &\textbf{Shadow}   &\textbf{Arrow} &\textbf{Dazzle} &\textbf{Curved} &\textbf{Total}\\ \midrule
C. Szegedy et a \cite{ResNet}    &\textcolor{blue}{87.4} &64.1 &60.6 &38.1 &60.7 &\textcolor{blue}{79.0} &54.1 &\textcolor{blue}{59.8} &66.7\\
Base model (ResNet-50)  &85.9 &64.5 &60.4 &38.9 &59.5 &76.2 &51.9 &58.9 &66.2\\
Base model+(A)  &85.7 &\textcolor{blue}{65.4} &61.7 &40.9 &59.1 &78.2 &53.2 &58.2 &66.9\\
Base model+(A)+(B)  &86.2 &64.3 &\textcolor{blue}{62.7} &\textcolor{blue}{41.9} &\textcolor{blue}{61.2} &77.3 &\textcolor{blue}{54.7} &58.5 &\textcolor{blue}{67.3}\\
\hline
C. Szegedy et al \cite{ResNet}     &90.2 &68.2 &65.9 &41.7 &64.6 &\textcolor{red}{84.0} &59.8 &\textcolor{red}{65.5} &70.8\\
Base model (ResNet-101) &88.3 &68.7 &64.2 &41.5 &66.4 &81.2 &60.3 &65.0 &68.2\\
Base model+(A) &89.9 &69.3 &\textcolor{red}{66.7} &42.2 &66.1 &80.8 &\textcolor{red}{61.7} &64.3 &70.3\\
Base model+(A)+(B) &\textcolor{red}{90.5} &\textcolor{red}{70.3} &66.2 &\textcolor{red}{43.9} &\textcolor{red}{67.2} &82.3 &58.8 &64.0
&\textcolor{red}{72.1}\\

\hline\bottomrule
\end{tabular}
\label{ablation}
\end{table*}

\begin{figure}
  \begin{minipage}{\linewidth}
  \includegraphics[width=.49\linewidth]{ 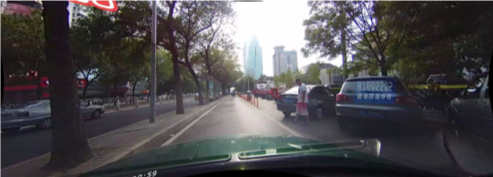}
  \includegraphics[width=.49\linewidth]{ 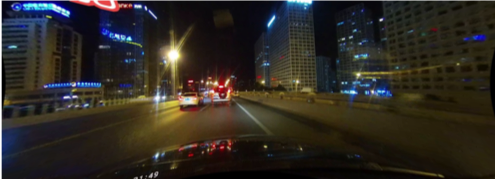}
  \end{minipage}%
  
  \begin{minipage}{\linewidth}
  \includegraphics[width=.49\linewidth]{ 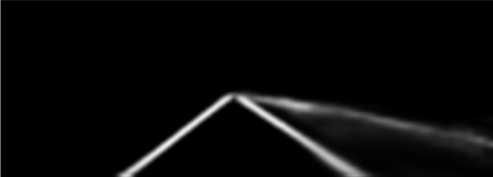}
  \includegraphics[width=.49\linewidth]{ 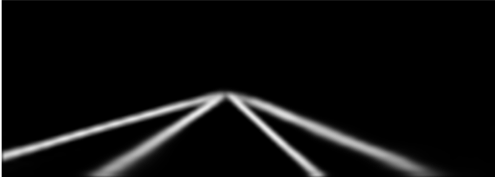}
  \end{minipage}%
  
  \begin{minipage}{\linewidth}
  \includegraphics[width=.49\linewidth]{ 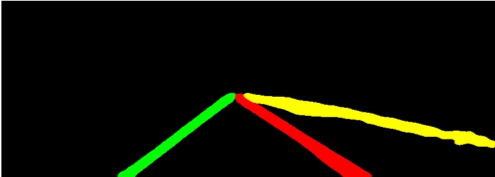}
  \includegraphics[width=.49\linewidth]{ 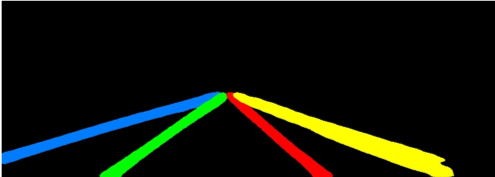}
  \end{minipage}%
  \caption{Illustration of ablation study. The first row is the input images, the second row is the rough segmentation results in branch (A) and the third row is the refined segmentation results after (B).}
  \label{figablation}
\end{figure}

\section{Ablation study and discussion}
We investigate the effect of all modules in the proposed framework including the base model, (A) the rough segmentation branch and (B) the refined segmentation branch to the segmentation results. Note that our base model shares the same structure as well as  experimental setting with \cite{ResNet} and thus achieves similar performance.

Reported by \cite{ResNet}, ResNet-101 \cite{ResNet} has better performance in lane segmentation than ResNet-50 since image feature is more fully learned with deeper neural network. In the Table \ref{ablation}, we can summarize that it is better to select network with effective image feature exploiting as the base model for the proposed ContinuityLearner because almost all frameworks based on ResNet-101 achieve higher accuracy than their correspondences based on ResNet-50. When the base model better comprehends image feature, it provides an relatively accurate rough segmentation with less false positive prediction (e.g. To recognise a background pixel as a lane pixel). Though (B) the refined segmentation branch is capable of predicting missing key points of lanes (false negative prediction), it is sensitive to the false positive in the coarse semantics outputted by (A). Consequentially, We also observe that reducing false positive prediction in rough segmentation results in higher refinement performance.

In Table \ref{ablation}, given the same base model, we analyze the contribution of branch (A) and branch (B) and obverse that both modules improve prediction accuracy in the average case. Branch (A) introduces global context as a constraint to exploit image only and brings unstable improvement among various traffic cases. For instance, based on ResNet-101, the branch (A) has lower accuracy in shadow, arrow and curved cases compared with the base model but it achieves better performance in the other cases. Compared with the coarse semantic outputted by (A), branch (B) has higher accuracy almost in every case, which verifies the geometric continuity feature is able to  represent shape knowledge of lanes more effectively than image feature. Fig. \ref{figablation} provides a clear illustration how branch (A) and branch (B) contributes to the final lane segmentation.

\section{Conclusion}
In this paper, we propose the ContinuityLearner, a novel deep-learning-based paradigm, that exploits geometric continuity feature representation conveying in traffic images and generates smooth semantic for lanes. The geometric continuity feature, as a compound of class-dependent image feature and pixel's spatial location, better represents objects that has weak textual coherency but strong geometric continuity in images. Specifically, the paradigm consists of a rough segmentation branch and a refined segmentation branch. In the rough segmentation branch, coarse semantic of lanes and class-dependent feature maps are calculated by a ResNet-based CNN with the context-encoding module. Fusing the coarse semantic and the class-dependent feature maps together, the geometric continuity feature is generated by an encoding layer. Finally, a MLP-based network performs on the geometric continuity feature and completes the course segmentation into a smooth and integrated semantic of lanes. Experiments over Tusimple and CULane datasets show that our proposed method delivers outstanding performance on lane instance segmentation and outperforms other existing methods, especially in scenes that lack visual information of lanes.

%
\IEEEpeerreviewmaketitle

{\small
\bibliographystyle{IEEEtran}
\bibliography{ContinuousNet}
}

\end{document}